\documentclass[sn-mathphys]{sn-jnl}

\usepackage{subcaption}
\usepackage{multirow}

\jyear{2025}%

\theoremstyle{thmstyleone}%
\theoremstyle{thmstyletwo}%

\theoremstyle{thmstylethree}%

\begin{document}

\title[Text2Cypher Across Languages: Evaluating and Finetuning LLMs]{Text2Cypher Across Languages: Evaluating and Finetuning LLMs}


\author*[1]{\fnm{Makbule Gulcin} \sur{Ozsoy}}\email{makbule.ozsoy@neo4j.com}

\author[1]{\fnm{William} \sur{Tai}}\email{william.tai@neo4j.com}


\affil*[1]{\orgname{Neo4j}, \orgaddress{\city{London}, \country{UK}}}
\affil[2]{Submitted to NLPIR 2025}



\abstract{
Recent advances in large language models (LLMs) have enabled natural language interfaces that translate user questions into database queries, such as Text2SQL, Text2SPARQL, and Text2Cypher. While these interfaces enhance database accessibility, most research today focuses on English, with limited evaluation in other languages. This paper investigates the performance of both foundational and finetuned LLMs on the Text2Cypher task across multiple languages. 
We create and release a multilingual dataset by translating English questions into Spanish and Turkish while preserving the original Cypher queries, enabling fair cross-lingual comparison. Using standardized prompts and metrics, we evaluate several foundational models and observe a consistent performance pattern: highest on English, followed by Spanish, and lowest on Turkish. We attribute this to differences in training data availability and linguistic features. We also examine the impact of translating task prompts into Spanish and Turkish. Results show little to no change in evaluation metrics, suggesting prompt translation has minor impact. 
Furthermore, we finetune a foundational model on two datasets: one in English only, and one multilingual. Finetuning on English improves overall accuracy but widens the performance gap between languages. In contrast, multilingual finetuning narrows the gap, resulting in more balanced performance. Our findings highlight the importance for multilingual evaluation and training to build more inclusive and robust query generation systems.
}




\maketitle

\section{Introduction}\label{sec1}

Databases provide efficient mechanisms for storing, organizing, and retrieving data. Query languages such as SQL (for relational databases), SPARQL (for RDF graphs), and Cypher (for graph databases) enable users to interact with these systems~\cite{hogan2021knowledge}. Recent advancements in large language models (LLMs) have enabled the translation of natural language questions into database queries (Text2SQL, Text2SPARQL, Text2Cypher), making databases more accessible to non-experts. However, most research in this area has focused on English, with limited investigation on multilingual capabilities~\cite{jannuzzi2024zero, geng2024not}.

\begin{figure}
  \centering
  \includegraphics[width=0.8\linewidth]{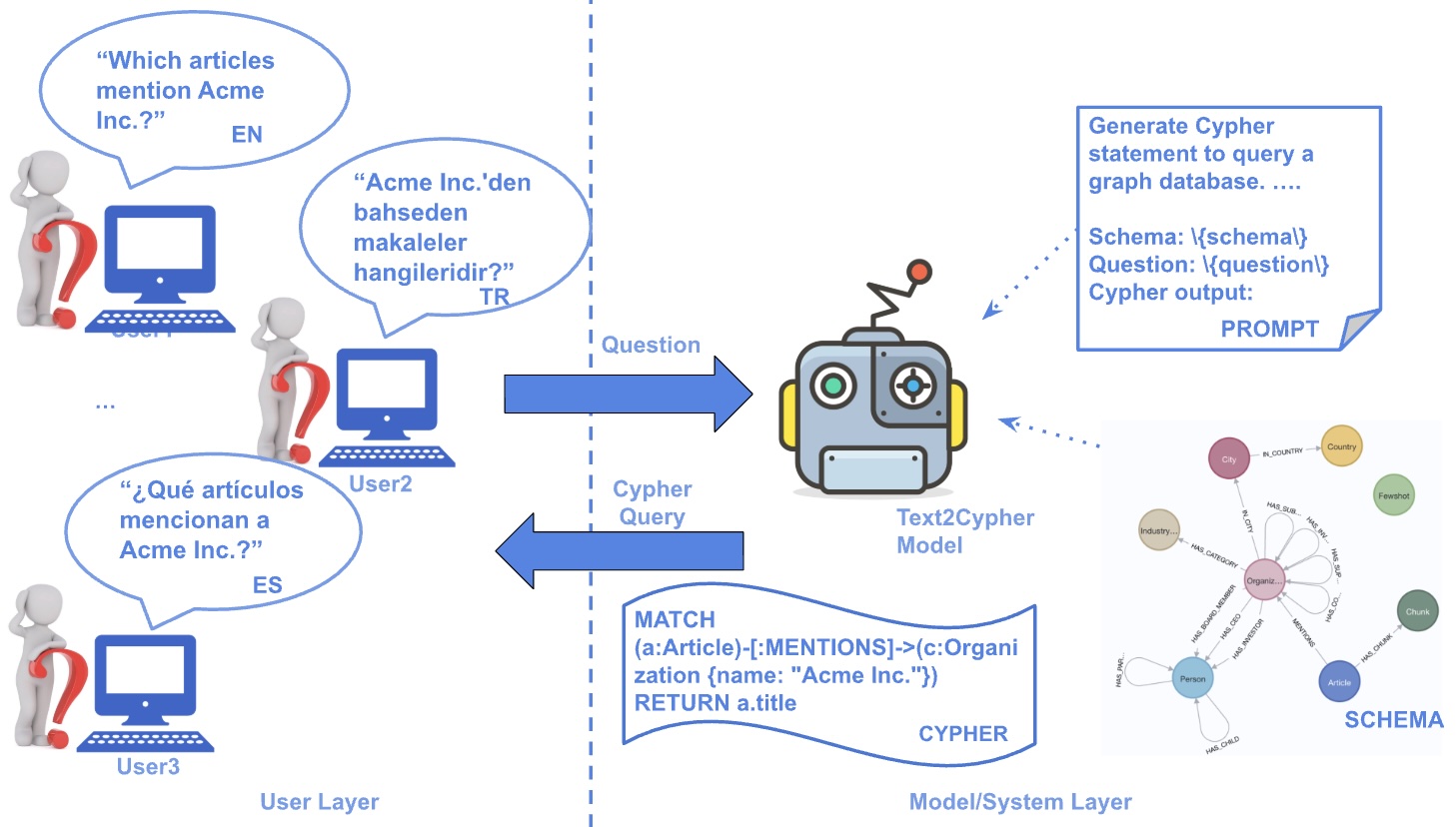}
    \caption{Multiple users access the same database and want to generate a Cypher query for the question “Which articles mention Acme Inc.?” However, they may express this question in different languages, such as English (EN), Spanish (ES), or Turkish (TR).
    }
  \label{fig:text2cypher}
\end{figure}

This work focuses on the multilingual Text2Cypher task, which involves converting natural language questions into executable Cypher queries (see Figure~\ref{fig:text2cypher}). For instance, users querying the same graph database might ask \texttt{"Which articles mention Acme Inc.?"} in English (EN), Spanish (ES), or Turkish (TR). A robust Text2Cypher system should generate the same query regardless of input language, such as: \texttt{MATCH (a:Article)-[:MENTIONS]->(c:Company \{name: "Acme Inc."\}) RETURN a.title}. 
We investigate how foundational and finetuned LLMs perform on this multilingual task. First, we create a multilingual dataset for training and evaluation in English, Spanish, and Turkish. Second, we examine whether translating the task instructions (i.e., prompts) into Spanish and Turkish affects model performance. Finally, we evaluate whether multilingual finetuning improves generalization by comparing models finetuned on English-only versus multilingual data.

Our main contributions are:
\begin{itemize}
\item We create and release a multilingual version of the Text2Cypher dataset~\cite{ozsoy2025text2cypher}, translated into Spanish and Turkish to support multilingual training and evaluation.\footnote{Test Set:\url{https://huggingface.co/datasets/mgoNeo4j/translated_text2cypher24_testset}}\footnote{Training Set: \url{https://huggingface.co/datasets/mgoNeo4j/translated_text2cypher24_trainset_sampled}}
\item We evaluate foundational LLMs on a multilingual test set, showing that performance is highest in English, followed by Spanish, and lowest in Turkish. We attribute the lower Turkish performance to linguistic differences and training data scarcity.
\item We analyze the effect of translating prompts into Spanish and Turkish and find minimal impact on performance in our setting.
\item We investigate the impact of finetuning. English-only finetuning increases accuracy but widens the performance gap across languages, while multilingual finetuning narrows the gap and yields more balanced performance.
\end{itemize}





The paper is organized as follows: Section~\ref{rel_work} reviews related work. Section~\ref{dataset} describes the multilingual dataset. Section~\ref{results} presents model evaluations. Section~\ref{conc} concludes the paper.

\section{Related Work} \label{rel_work}
This section reviews related work on large language models (LLMs) for non-English content, as well as multilingual approaches to database query generation tasks such as Text2SQL and Text2SPARQL.

\subsection{LLMs and Non-English Content}
Large language models (LLMs) have become central to a wide range of natural language processing tasks. While many early LLMs were trained primarily on English data, their capabilities have recently been extended to other languages through the development of multilingual models~\cite{nicholas2023lost}. Lai et al.~\cite{lai2024llms} describe the multilingual capabilities of LLMs in terms of their ability to handle task instructions, inputs, and outputs in different languages. They identify two multilingual capabilities: (i) Understanding, where the model interprets instructions in various languages and generates a correct output, and (ii) Generating, where a fixed language (e.g., English) is used for the instruction and the model produces outputs in different target languages. Both capabilities depend heavily on the model’s training data and methodology.

In order to support and enhance multilingual capabilities, various approaches have been proposed in the literature. Some studies leverage multilingual parallel data to continue pretraining~\cite{yang2023bigtranslate, zhu2023extrapolating}. Others employ multilingual instruction data for fine-tuning~\cite{ustun2024aya, luo2023yayi, li2023bactrian, lai2023okapi}. A third line of work focuses on cross-lingual prompting during inference~\cite{huang2023not, etxaniz2023multilingual}. 
Researchers have also begun exploring the internal mechanisms of these models~\cite{zhao2024llama, kargaran2024mexa, zhong2024beyond, schut2025multilingual, bandarkar2024layer} and evaluating their multilingual performance across various application domains~\cite{ozsoy2024multilingual, jannuzzi2024zero}. 

Multilingual models are typically trained on datasets that include multiple languages, allowing them to learn cross-lingual representations. This enables models to transfer word associations and grammatical patterns learned from high-resource languages, such as English, to languages with less available training data. However, since English still dominates most training corpora, these models often internalize linguistic patterns and assumptions specific to English~\cite{nicholas2023lost, zhao2024llama}. As a result, LLMs tend to perform significantly better in higher-resource languages and those similar to them, compared to lower-resource languages~\cite{nicholas2023lost}.

\subsection{Multilingual Database Query Generation}

Much of the multilingual work in translating natural language to database query languages has focused on the Text2SQL task~\cite{dou2023multispider, jose2021mrat,huang2025exploring}. Many studies translate the English-language Spider dataset~\cite{yu2018spider} into other languages, such as Chinese in CSpider~\cite{min2019pilot}, Turkish in TURSpider~\cite{kanburoglu2024turspider}, Arabic in Ar-Spider~\cite{almohaimeed2024ar}, and seven languages in MultiSpider~\cite{dou2023multispider}. Other works have created new multilingual datasets using human annotators, such as StatBot.Swiss~\cite{nooralahzadeh2024statbot}, which includes data in English and German. Additional languages like Portuguese~\cite{jannuzzi2024zero, pedroso2025performance} and Russian~\cite{bakshandaeva2022pauq} have also been explored. Jannuzzi et al.~\cite{jannuzzi2024zero} evaluate Text2SQL performance in Portuguese using zero-shot prompting and report a performance gap between using English and Portuguese prompts.

For SPARQL, although there are multilingual question answering (QA) datasets and methods~\cite{cui-etal-2022-compositional, srivastava2024mst5, perevalov2024multilingual}, they are not specifically designed for the Text2SPARQL task. Recently, the Text2SPARQL Challenge~\cite{text2sparqlChallenge25} released a dataset containing questions in English and Spanish. Additionally, Perevalov et al.~\cite{perevalov2024towards} explored the reverse task, translating SPARQL to natural language, using English, German, and Russian in their evaluation. Recent work has also investigated the use of LLMs and SPARQL queries for knowledge graph reasoning, such as LLM4QA~\cite{lan2024llm4qa}, which highlights the broader potential of query generation tasks across different database languages.

In this work, we study the Text2Cypher task, which translates natural language questions into Cypher queries.

\section{Multilingual Dataset Preparation} \label{dataset}
As the base dataset, we used the publicly available Text2Cypher dataset~\cite{ozsoy2025text2cypher}, which includes English questions, database schema, ground-truth Cypher queries, and metadata. In order to evaluate foundational and finetuned models across languages, we created a multilingual dataset by translating the original English questions into Spanish and Turkish. 

\subsection{Text2Cypher Beyond English}
We have chosen English, Spanish and Turkish for the comparisons based on their variations in their resourcedness levels, language families and linguistic characteristics:
\begin{itemize}
    \item \textbf{Language resourcedness:} This indicates availability of data, tools, and research support for a language. It impacts LLM performance, where models generally perform better on languages with richer and more diverse training data~\cite{nicholas2023lost}. While English is an extremely high-resource language, Spanish is a high-resource, and Turkish is a medium-resource language~\cite{nicholas2023lost,joshi2020state}. 
    
    \item \textbf{Language families:} This indicates relationships between languages, grouping them based on common origins and shared linguistic features. According to Dhamecha et al.~\cite{dhamecha2021role}, models generalize more effectively across linguistically similar languages. English and Spanish both belong to the Indo-European family, whereas Turkish belongs to the Altaic family, with distinct linguistic traits. 
    
    \item \textbf{Linguistic characteristics:} Turkish is an agglutinative language, where words are formed by concatenating multiple suffixes encoding grammatical functions, such as tense, case, person. This morphological richness leads to challenges in tokenization and semantic parsing, and increases data requirements for LLMs~\cite{bayram2024setting, alecakir2022turkishdelightnlp}. 
    Additionally, both Spanish and Turkish allow relatively flexible word order compared to English and are pro-drop languages, allowing omission of subjects and pronouns. 
    These characteristics increase ambiguity and makes the mapping to structured queries more difficult compared to English~\cite{bayram2024setting, alecakir2022turkishdelightnlp, tohma2020challenges, baucells2025iberobench}. 
\end{itemize}
Given these factors, we expect a moderate performance drop for Spanish and a more substantial drop for Turkish due to its linguistic divergence.

\subsection{Multilingual Test Set} \label{test_set}
In order to enable multilingual evaluation, we automatically translated the original English questions from the Text2Cypher dataset~\cite{ozsoy2025text2cypher} into Spanish and Turkish using an LLM. We used the following steps for translation:
\begin{itemize}
    \item \textbf{Masking Named Entities and Quotes}: To ensure that named entities and quoted text remained unchanged during translation, we first masked them. Named entities were replaced with placeholders indicating their type and index (e.g., \texttt{LOCATION\_0}), while quoted strings were replaced with \texttt{QUOTE\_<index>} markers. For example, the sentence \texttt{"Hello, I work at 'Neo4j' in London"} was masked as \texttt{"Hello, I work at QUOTE\_0 in LOCATION\_0"}. 
    
    \item \textbf{Translation Using an LLM}: We translated the masked questions using the GPT-4o-mini model, guided by the prompt shown in Table~\ref{tab:translate_prompt}. 
    
    \item \textbf{Restoring Masks}: After translation, we replaced the placeholders with their original named entities and quoted expressions to reconstruct the final question.
\end{itemize}

This process produced semantically aligned triples across the three languages. During translation, 50 samples with masking-related issues were excluded from the test set. Consequently, the final test set contains 4,783 samples per language. 
Translation quality was evaluated using COMET-KIWI-22~\cite{rei2022cometkiwi}, a reference-free metric ranging from 0 to 1, where higher values indicate better quality. Our test dataset obtained $0.7934$ for Turkish and $0.8062$ for Spanish translations. 
Note that the ground-truth Cypher queries were not translated. These queries contain (i) Cypher-specific syntax (e.g., \texttt{MATCH, WHERE}), (ii) terms from the schema (e.g., \texttt{Person, ACTED\_IN, Movie}), and (iii) user-provided literals (e.g., \texttt{"Tom Hanks"}), all of which are intended to remain in their original form. 

\begin{table} %
  \caption{Translation prompt for \{language\_name\}}
  \label{tab:translate_prompt}
  \begin{tabular}{p{0.9\textwidth}}
    \toprule
    \textbf{Translation instruction prompt}  \\
    \midrule
    You are a professional translator. Your task is to translate the given text from English to \{language\_name\}. 
    Follow these guidelines:\\
    1. Maintain the original meaning and tone of the text \\
    2. Preserve any placeholders in the format \verb|<TYPE_NUMBER>| (e.g., \verb|<PERSON_0>|, \verb|<QUOTE_1>|) \\
    3. Keep the translation natural and fluent in \{language\_name\}\\
    4. Maintain proper formatting and punctuation\\
    5. Do not translate or modify any placeholders\\
    6. Do not translate any text within quotation marks\\
    7. Do not translate named entities (proper nouns, names, places, organizations)\\
    8. Do not translate numbers, dates, or measurements\\
    Please provide only the translated text without any explanations or additional context. \\
    \bottomrule
\end{tabular}
\end{table}

\subsection{Multilingual Training Set} \label{training_set}
The training set is based on the original English Text2Cypher training dataset~\cite{ozsoy2025text2cypher}, which contains approximately 40,000 instances. To create a multilingual training dataset, we prepared around 36,000 instances in total, distributed evenly across English(EN), Spanish(ES), and Turkish(TR); such that about 12,000 instances per language. These instances are randomly sampled from the original data and only the user question field is translated, while the Cypher queries and other fields are kept unchanged. The question translation process followed the same method described for the test set in Section~\ref{test_set}. The resulting training dataset includes: 
(i) Around 6,750 questions common to all three languages,
(ii) Around 1,500 questions shared between each pair of languages (i.e., EN-ES, EN-TR, ES-TR),
(iii) Approximately 3,800 unique questions per language. 
This structure provides fully aligned questions for direct cross-lingual learning and evaluation, alongside unique, language-specific questions to capture linguistic diversity. 
For the training set, COMET-KIWI-22~\cite{rei2022cometkiwi} was also applied to evaluate the translation quality. This is a reference-free metric scores in the range $[0,1]$, where higher values reflecting better quality. Our multilingual training dataset obtained $0.7849$ for Turkish and $0.7999$ for Spanish translations.

We also created an English-only training set alongside the multilingual dataset to enable comparison with models finetuned exclusively on English data. For this, we selected all unique instance ids (removing duplicates across languages) and extracted the corresponding English questions from the original dataset. This resulted in 20,512 unique questions. This setup allows a fair comparison between finetuning on English-only versus multilingual data with similar data coverage.

\section{Experimental Results} \label{results}

In this section, we evaluate foundational LLMs on the Text2Cypher task across English, Spanish, and Turkish. We then compare their performance to models finetuned on English-only and multilingual training datasets. 

\begin{table}
\caption{Instructions used for Text2Cypher task}
  \label{tab:instructions}
  \begin{tabular}{p{0.15\linewidth}p{0.75\linewidth}}
    \toprule
    \textbf{Type} & \textbf{Instruction prompt}  \\
    \midrule
    System \newline Instruct. &  Task: Generate Cypher statement to query a graph database. Instructions: Use only the provided relationship types and properties in the schema. Do not use any other relationship types or properties that are not provided in the schema. Do not include any explanations or apologies in your responses. Do not respond to any questions that might ask anything else than for you to construct a Cypher statement. Do not include any text except the generated Cypher statement. \\
    \hline
    User \newline Instruct. & Generate Cypher statement to query a graph database. Use only the provided relationship types and properties in the schema. \newline
            Schema: \{schema\} \textbackslash n
            Question: \{question\} \textbackslash n
            Cypher output: 
         \\
  \bottomrule
\end{tabular}
\end{table}

\subsection{Experimental Setup}
We conduct our experiments on the test split of the newly created multilingual Text2Cypher dataset containing 4,783 samples for each language, which are translated from the original English questions into Spanish and Turkish. 
We use the following foundational models:
(i) \textbf{Gemma-2-9b-it:} Released in June 2024. Primarily supports English. 
(ii) \textbf{Meta-Llama-3.1-8B-Instruct:} Released in July 2024. Primarily supports English and several European languages. 
(iii) \textbf{Qwen2.5-7B-Instruct:} Released in September 2024. Primarily supports Chinese and English and maintains multilingual support for over 29 languages. 
We use their Unsloth versions of models, finetuned with Instruct and quantized to 4-bit precision (e.g., unsloth/gemma-2-9b-it-bnb-4bit) for efficiency and ease of use. 
For the Text2Cypher task, we use the same prompts as prior work~\cite{ozsoy2025text2cypher}, shown in Table~\ref{tab:instructions}. 
After generation, a post-processing step is used for removing unwanted text, such as the 'cypher:' suffix. 

We use the HuggingFace Evaluate library~\cite{hfEvaluate} to compute evaluation metrics. We employ two evaluation procedures:
(i) \textbf{Translation-based:} Compares generated Cypher queries with reference queries based on their textual content. We report the ROUGE-L score for this evaluation.
(ii) \textbf{Execution-based:} Executes both the generated and reference queries on the target database and compares their outputs (sorted lexicographically). 
Since only a subset of the dataset (approximately 50\%) has active database access, execution-based evaluation is conducted on this subset only. 
We report the Exact-Match score for this evaluation.

\begin{figure*}
\centering
    \begin{subfigure}[b]{0.49\linewidth}
        \includegraphics[trim=10 0 20 0, clip, width=\linewidth]{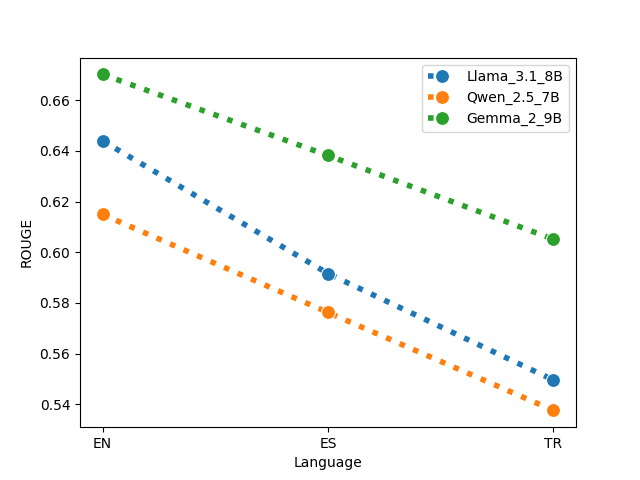}
        \caption{Translation-based: ROUGE-L score}
    \end{subfigure}
    \hfill
    \begin{subfigure}[b]{0.49\linewidth}
        \includegraphics[trim=10 0 20 0, clip, width=\linewidth]{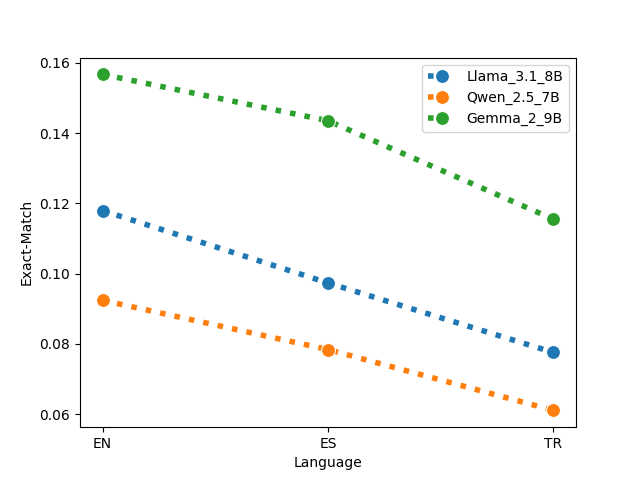}
        \caption{Execution-based: Exact-Match score}
    \end{subfigure}

    \caption{Performance of models on Text2Cypher when input questions are in English (EN), Spanish (ES) or Turkish (TR).}
    \label{fig:text2cypher_results}
\end{figure*}

\subsection{Impact of question language}
We first examine the impact of question language on foundational LLMs using the translated dataset. In this stage, we do not modify the task prompt for the Text2Cypher task (shown in Table~\ref{tab:instructions}), which is kept in English. 
Figure~\ref{fig:text2cypher_results} shows the performance of the foundational models on English, Spanish, and Turkish questions.

The results indicate that all models perform best on English questions, followed by Spanish, with the lowest performance observed for Turkish. This pattern aligns with differences in language resourcedness of these languages and underlines the effect of training data coverage. LLMs generally perform better on languages with richer and more diverse training data, such as English, which improves the quality of learned representations~\cite{nicholas2023lost,joshi2020state}. 
Additionally, prior work~\cite{dhamecha2021role} shows that  models generalize more effectively across linguistically similar languages. Spanish and English both belong to the Indo-European language family, while Turkish belongs to the Altaic family, and exhibits distinct linguistic traits. 
Languages with complex morphology and flexible word order, such as Turkish, have greater challenges for semantic parsing and alignment to Cypher queries, as discussed above. Spanish, being linguistically closer to English in structure, experiences less from these challenges. Overall, these characteristics increase ambiguity and makes the mapping to structured queries more difficult compared to English~\cite{bayram2024setting, alecakir2022turkishdelightnlp, tohma2020challenges, baucells2025iberobench}. 

In these experiments, the prompts remained in English, which may introduce additional challenges for LLMs when processing non-English questions. In the next section, we explore this effect further by evaluating model performance using both translated questions and translated prompts.

\begin{table}
\caption{Spanish (ES) and Turkish (TR) Instructions used for Text2Cypher}
  \label{tab:instructionsESTR}
  \begin{tabular}{p{0.07\linewidth}p{0.10\linewidth}p{0.70\linewidth}}
    \toprule
    \textbf{Lang.} &  \textbf{Type} & \textbf{Instruction prompt}  \\
    \midrule
    \multirow[c]{2}{*}{ES} & System \newline Instruct. &  Tarea: Generar una sentencia Cypher para consultar una base de datos de grafos. Instrucciones: Usa únicamente los tipos de relaciones y propiedades proporcionados en el esquema. No utilices ningún otro tipo de relación ni propiedades que no estén incluidas en el esquema. No incluyas explicaciones ni disculpas en tus respuestas. No respondas a ninguna pregunta que no sea una solicitud para construir una sentencia Cypher. No incluyas ningún texto excepto la sentencia Cypher generada. \\
    \noalign{\vskip 0.7em}
     & User \newline Instruct. & Genera una sentencia Cypher para consultar una base de datos de grafos. Usa únicamente los tipos de relaciones y propiedades proporcionados en el esquema. \newline 
        Esquema: \{schema\} \textbackslash n
        Pregunta: \{question\} \textbackslash n
        Salida Cypher:
        \\
    \midrule
    \multirow[c]{2}{*}{TR} & System \newline Instruct. &  Görev: Bir çizge veritabanını sorgulamak için bir Cypher ifadesi oluştur. 
    Talimatlar: Yalnızca şemada verilen ilişki türlerini ve özellikleri kullan. Şemada verilmeyen herhangi bir ilişki türünü veya özelliği kullanma. Yanıtlarında hiçbir açıklama veya özür ifadesine yer verme. Cypher ifadesi oluşturmak dışında başka bir şey isteyen sorulara yanıt verme. Oluşturulan Cypher ifadesi dışında hiçbir metin ekleme.\\

    \noalign{\vskip 0.7em} 
    & User \newline Instruct. & Bir çizge veritabanını sorgulamak için Cypher ifadesi oluştur. Şemada verilen ilişki türleri ve özellikler dışında hiçbir şeyi kullanma. \newline
        Şema: \{schema\} \textbackslash n
        Soru: \{question\} \textbackslash n
        Cypher çıktısı: 
         \\
    
    \bottomrule
\end{tabular}
\end{table}

\begin{figure*}
\centering
    \begin{subfigure}[b]{0.45\linewidth}
        \includegraphics[trim=10 0 20 0, clip, width=\linewidth]{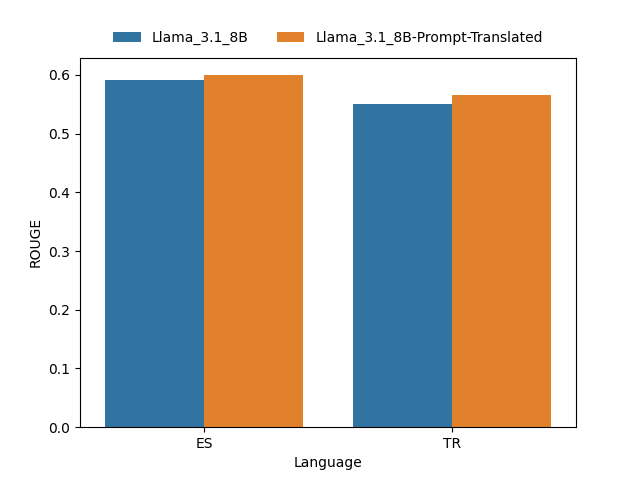}
        \caption{Translation-based: ROUGE-L score}
    \end{subfigure}
    \hfill
    \begin{subfigure}[b]{0.45\linewidth}
        \includegraphics[trim=10 0 20 0, clip, width=\linewidth]{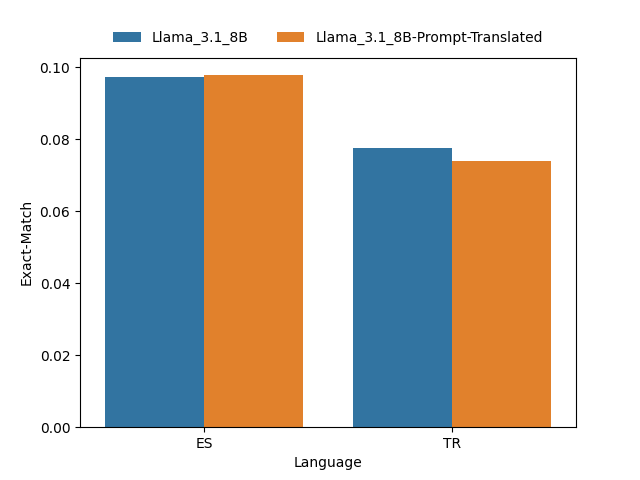}
        \caption{Execution-based: Exact-Match score}
    \end{subfigure}

    \caption{Model performance with/out prompt translation.}
    \label{fig:text2cypher_results_afterprompt}
\end{figure*}

\subsection{Impact of prompt language}
We also translated the prompts into Spanish and Turkish to evaluate the impact of prompt language on model performance. The prompts were translated using the GPT-4o-mini model and are shown in Table~\ref{tab:instructionsESTR}. The performance comparison for Spanish and Turkish with and without translated prompts are presented in Figure~\ref{fig:text2cypher_results_afterprompt}. 
The results show that when prompts are translated, the ROUGE-L scores, which used for translation-based evaluation comparing ground-truth and generated text, improve slightly (by about 1–1.5\%). However, Exact-Match scores for execution-based evaluation remain largely unchanged. These results suggest that prompt translation has a minor impact on the overall result in our setup. 

It is important to note that in our dataset, the database schemas, including nodes, relationships and properties, remain in English. Future work could explore the impact of localizing schema items to match the language of the input and prompt. 

\subsection{Impact of finetuning}

After observing performance differences across languages, we next explore how finetuning affects results. Among the evaluated models, we selected Meta-Llama-3.1-8B-Instruct for fine-tuning due to its multilingual support, open availability, and compatibility with our Unsloth-based setup. Considering the time and resource constraints, we focused on a single model and considered fine-tuning others as a future work.  
We finetune the selected baseline model separately on two datasets: 
\begin{itemize}
\item English-only finetuned model: Trained on the English-only dataset described in Section~\ref{training_set}. This setup allows us to evaluate how English-only finetuning affects performance on both English and non-English inputs.
\item Multilingual finetuned model: Trained on the multilingual dataset to assess the benefits of exposure to multiple languages during finetuning.
\end{itemize} 

\begin{figure*}
\centering
    \begin{subfigure}[b]{0.49\textwidth}
        \includegraphics[trim=10 0 20 0, clip, width=\linewidth]{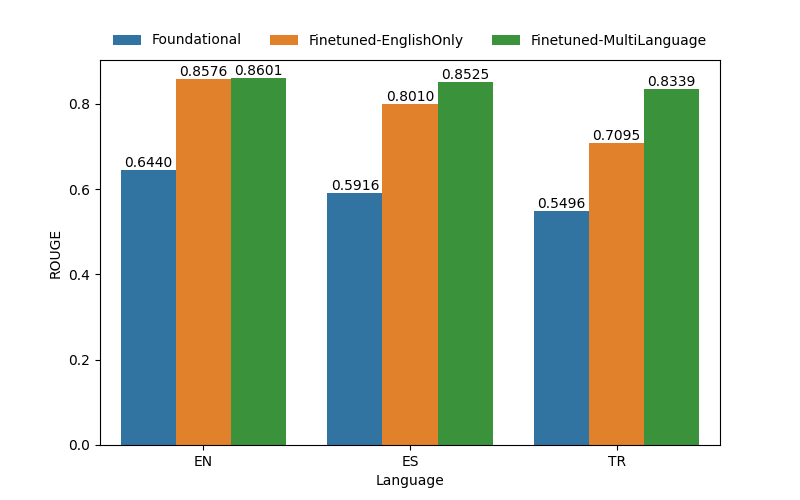}
        \caption{Translation-based: ROUGE-L score}
    \end{subfigure}
    \hfill
    \begin{subfigure}[b]{0.49\textwidth}
        \includegraphics[trim=10 0 20 0, clip, width=\linewidth]{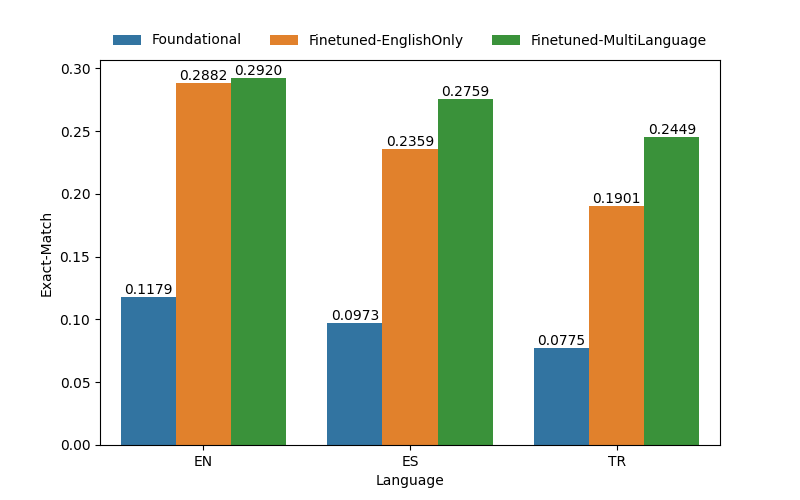}
        \caption{Execution-based: Exact-Match score}
    \end{subfigure}

    \caption{Comparison of model performance: Foundational model (blue), Fine-tuned model on English-only data (orange), and Fine-tuned model on multilingual data (green).}
    \label{fig:text2cypher_results_afterfinetune}
\end{figure*}

The finetuned models are evaluated on the test set using data per language. In Figure \ref{fig:text2cypher_results_afterfinetune} we present the performance of the foundational model, finetuned model on English-only dataset and finetuned model on multilingual dataset. 
The figure shows that using the English-only dataset improves model performance of the Text2Cypher task across all languages. However, the improvement for Turkish is smaller compared to English and Spanish. For example, based on the translation-based ROUGE-L score, English and Spanish improve by around 0.20, while Turkish improves by approximately 0.15. 
Spanish still performs worse than English, followed by Turkish. This performance gap may be attributed to linguistic differences: English and Spanish belong to the Indo-European family, whereas Turkish is an Altaic language with different characteristics. 
When evaluating the model finetuned on the multilingual dataset, we observe that performance across languages becomes more balanced. For instance, ROUGE-L scores reach 0.86 for English, 0.85 for Spanish, and 0.83 for Turkish. These results suggest that multilingual finetuning improves generalization and narrows the performance gap across languages.

Overall, these results indicate that both language resourcedness and linguistic characteristics influence model performance. Foundational models and those finetuned only on English struggle with non-English inputs, particularly for languages with different characteristics or with less resources. Multilingual finetuning mitigates these challenges and leads to more balanced performance across languages.

\section{Conclusion} \label{conc}
This work presents a performance comparison of foundational and finetuned LLMs on the Text2Cypher task across English, Spanish, and Turkish. We created and released a multilingual dataset by translating user questions in English into Spanish and Turkish, enabling fair cross-lingual evaluation. 
Our results consistently show that foundational models perform best on English inputs, followed by Spanish, and then Turkish. This pattern is likely driven by differences in language resourcedness and linguistic characteristics, which introduce additional challenges for LLMs. We also explored the effect of translating task instruction prompts alongside the input questions. Our findings suggest prompt language has minimal impact on overall performance in our setting. 
Finally, we finetuned a foundational model on both English-only and multilingual datasets. The results revealed that English-only finetuning improves performance across all languages, it does not eliminate the performance gap. In contrast, multilingual finetuning improves generalization and narrows the performance gap across languages. 
Future directions include schema localization to further support multilingual performance, expanding evaluation to additional languages, and exploring alternative alignment strategies beyond finetuning to improve cross-lingual robustness.

\begin{appendices}
\section*{Declaration on Generative AI Usage}

  
During the preparation of this work, the author(s) used Chat-GPT in order to: 'Improve writing style' and 'Paraphrase and reword'. After using these tool(s)/service(s), the author(s) reviewed and edited the content as needed and take(s) full responsibility for the publication’s content. 






\end{appendices}


\bibliography{main-sn-article}


\end{document}